%% file: main.tex
\pdfoutput=1
\documentclass[mathpazo]{cicp}

\usepackage{microtype}
\usepackage{graphicx}
\usepackage{grffile}
\usepackage{subfigure}
\usepackage{booktabs} 
\usepackage{url}
\usepackage[utf8]{inputenc} 
\usepackage[T1]{fontenc}    
\usepackage{booktabs}       
\usepackage{amsfonts}       
\usepackage{nicefrac}       
\usepackage{microtype}      
\usepackage[ruled,linesnumbered,vlined]{algorithm2e}
\usepackage{amsmath}
\usepackage{xcolor}
\usepackage{enumitem}
\usepackage{graphicx}
\usepackage{amsthm}
\usepackage{multirow}
\usepackage{diagbox}
\usepackage{appendix}
\usepackage{ulem}
\usepackage{pifont}
\usepackage{amssymb}
\newcommand{\cmark}{\ding{51}}%
\newcommand{\xmark}{\ding{55}}%
\def\semichecked{\checkmark\!\!\!\raisebox{0.4 em}{\tiny$\smallsetminus$}}
\newcommand{\citep}[1]{\cite{#1}}
\usepackage{comment}
\newcommand{\para}[1]{\noindent\paragraph{#1}}

\begin{document}
\title{Learning to Discretize: Solving 1D Scalar Conservation Laws via Deep Reinforcement Learning}


\author[Wang Y F et.~al.]{Yufei Wang\affil{1}
\equalcontribution,
      Ziju Shen\affil{2}
      $^\dagger$, 
      Zichao Long\affil{3} and Bin Dong\affil{4}\comma\corrauth}
\address{\affilnum{1}\ Computer Science Department,
Carnegie Mellon University, USA. \\
          \affilnum{2}\ Academy for Advanced Interdisciplinary Studies, Peking University, P.R. China.\\
          \affilnum{3}\ School of Mathematical Sciences, Peking University, P.R. China.\\
          \affilnum{4}\ Beijing International Center for Mathematical Research, and Institute for Artificial Intelligence, Peking University, P.R. China.}
\emails{{\tt yufeiw2@andrew.cmu.edu} (Y. F.~Wang), {\tt zjshen@pku.edu.cn} (Z. J.~Shen),
        {\tt zlong@pku.edu.cn} (Z. C.~Long),{\tt dongbin@math.pku.edu.cn} (B. Dong)}


\begin{abstract}
Conservation laws are considered to be fundamental laws of nature. It has broad applications in many fields, including physics, chemistry, biology, geology, and engineering. Solving the differential equations associated with conservation laws is a major branch in computational mathematics. The recent success of machine learning, especially deep learning in areas such as computer vision and natural language processing, has attracted a lot of attention from the community of computational mathematics and inspired many intriguing works in combining machine learning with traditional methods. In this paper, we are the first to view numerical PDE solvers as an MDP and to use (deep) RL to learn new solvers.
As proof of concept, we focus on 1-dimensional scalar conservation laws. We deploy the machinery of deep reinforcement learning to train a policy network that can decide on how the numerical solutions should be approximated in a sequential and spatial-temporal adaptive manner. We will show that the problem of solving conservation laws can be naturally viewed as a sequential decision-making process, and the numerical schemes learned in such a way can easily enforce long-term accuracy. 
Furthermore, the learned policy network is carefully designed to determine a good local discrete approximation based on the current state of the solution, which essentially makes the proposed method a meta-learning approach. In other words, the proposed method is capable of learning how to discretize for a given situation mimicking human experts. Finally, we will provide details on how the policy network is trained, how well it performs compared with some state-of-the-art numerical solvers such as WENO schemes, and supervised learning based approach L3D and PINN, and how well it generalizes. 
\end{abstract}


\maketitle



\input{01_intro.tex}

\input{02_preliminaries.tex}
\input{03_algos.tex}
\input{04_experiments.tex}

\input{05_conclusion.tex}

\bibliographystyle{plain}
\bibliography{00_bibliography}
\newpage

\end{document}

%% file: 01_intro.tex
\section{Introduction}
Conservation laws are considered to be one of the fundamental laws of nature, and has broad applications in multiple fields such as physics, chemistry, biology, geology, and engineering. For example, Burgers equation, a very classic partial differential equation (PDE) in conservation laws, has important applications in fluid mechanics, nonlinear acoustics, gas dynamics, and traffic flow.

Solving the differential equations associated with conservation laws has been a major branch of computational mathematics \citep{leveque1992numerical,leveque2002finite}, and a lot of effective methods have been proposed, from classic methods such as the upwind scheme, the Lax-Friedrichs scheme, to the advanced ones such as the ENO/WENO schemes \citep{liu1994weighted, shu1998essentially}, the flux-limiter methods \citep{jerez2010new}, and etc. In the past few decades, these traditional methods have been proven successful in solving conservation laws. Nonetheless, the design of some of the high-end methods heavily relies on expert knowledge and the coding of these methods can be a laborious process. To ease the usage and potentially improve these traditional algorithms, machine learning, especially deep learning, has been recently incorporated into this field. For example, the ENO scheme requires lots of `if/else' logical judgments when used to solve complicated system of equations or high-dimensional equations. This very much resembles the old-fashioned expert systems. The recent trend in artificial intelligence (AI) is to replace the expert systems by the so-called `connectionism', e.g., deep neural networks, which leads to the recent bloom of AI. Therefore, it is natural and potentially beneficial to introduce deep learning in traditional numerical solvers of conservation laws.

\subsection{Related works}
In the last few years, neural networks (NNs) have been applied to solving ODEs/PDEs or the associated inverse problems. These works can be roughly classified into two categories according to the way that the NN is used.

The first type of works propose to harness the representation power of NNs, and are irrelevant to the numerical discretization based methods. For example, in the pioneering works \cite{weinan2018deep,sirignano2018dgm,raissi2017physics,raissi2019physics,bar2018data}, NNs are used as new ansatz to approximate solutions of PDEs. It was later generalized by \cite{wei2019general} to allow randomness in the solution which is trained using policy gradient. More recent works along this line include \citep{magiera2019constraint,michoski2019solving,both2019deepmod}. 
Besides, several works have focused on using NNs to establish direct mappings between the parameters of the PDEs  (e.g. the coefficient field or the ground state energy) and their associated solutions \citep{khoo2017solving,khoo2019switchnet,li2019variational,fan2018multiscale}. Furthermore, \cite{han2018solving,beck2017machine} proposed a method to solve very high-dimensional PDEs by converting the PDE to a stochastic control problem and use NNs to approximate the gradient of the solution.
The second type of works, which target at using NNs to learn new numerical schemes, are closely related to our work. However, we note that these works mainly fall in the setting of supervised learning (SL). For example, \cite{discacciati2020controlling} proposed to integrate NNs into high-order numerical solvers to predict artificial viscosity; \cite{ray2018artificial} trained a multilayer perceptron to replace traditional indicators for identifying troubled-cells in high-resolution schemes for conservation laws. 
These works greatly advanced the development in machine learning based design of numerical schemes for conservation laws. Note that in \cite{discacciati2020controlling}, the authors only utilized the one-step error to train the artificial viscosity networks without taking into account the long-term accuracy of the learned numerical scheme. \cite{ray2018artificial} first constructed several functions with known regularities and then used them to train a neural network to predict the location of discontinuity, which was later used to choose a proper slope limiter. Therefore, the training of the NNs is separated from the numerical scheme. Then, a natural question is whether we can learn discretization of differential equations in an end-to-end fashion and the learned discrete scheme also takes long-term accuracy into account. This motivates us to employ reinforcement learning to learn good solvers for conservation laws.



\subsection{Our Approach}
The main objective of this paper is to design new numerical schemes in an autonomous way. We propose to view iterative numerical schemes solving the conservation laws as a Markov Decision Process (MDP) and use reinforcement learning (RL) to find new and potentially better data-driven solvers. We carefully design an RL-based method so that the learned policy can generate high accuracy numerical schemes and can well generalize in varied situations. Details will be given in section \ref{sec::methods}.

Here, we first summarize the benefits and potentials of using RL to solve conservation laws (the arguments may also well apply to other evolution PDEs):

\begin{itemize}[leftmargin=*]
\item Most of the numerical solvers of conservation law can be interpreted naturally as a sequential decision making process (e.g., the approximated grid values at the current time instance definitely affects all the future approximations). Thus, it can be easily formulated as a MDP and solved by RL. 

\item In almost all the RL algorithms, the policy $\pi$ (which is the AI agent who decides on how the solution should be approximated locally) is optimized with regards to the values $Q^{\pi}(s_0, a_0) = r(s_0, a_0) + \sum_{t = 1}^\infty \gamma^t r(s_t, a_t)$, which by definition considers the long-term accumulated reward (or, error of the learned numerical scheme), thus could naturally guarantee the long-term accuracy of the learned schemes, instead of greedily deciding the local approximation which is the case for most numerical PDEs solvers. Furthermore, it can gracefully handle the cases when the action space is discrete, which is in fact one of the major strength of RL.

\item By optimizing towards long-term accuracy and effective exploration, we believe that RL has a good potential in improving traditional numerical schemes, especially in parts where no clear design principles exist. For example, although the WENO-5 scheme achieves optimal order of accuracy at smooth regions of the solution \citep{shu1998essentially}, the best way of choosing templates near singularities remains unknown. Our belief that RL could shed lights on such parts is later verified in the experiments: the trained RL policy demonstrated new behaviours and is able to select better templates than WENO and hence approximate the solution better than WENO near singularities.

\item Non-smooth norms such as the infinity norm of the error is often used to evaluate the performance of the learned numerical schemes. As the norm of the error serves as the loss function for the learning algorithms, computing the gradient of the infinity norm can be problematic for supervised learning, while RL does not have such problem since it does not explicitly take gradients of the loss function (i.e. the reward function for RL).

\item  Learning the policy $\pi$ within the RL framework makes the algorithm meta-learning-like \cite{schmidhuber1987evolutionary,bengio1992optimization,andrychowicz2016learning,li2016learning,finn2017model}. The learned policy $\pi$ can decide on which local numerical approximation to use by judging from the current state of the solution (e.g. local smoothness, oscillatory patterns, dissipation, etc). This is vastly different from regular (non-meta-) learning where the algorithms directly make inference on the numerical schemes without the aid of an additional network such as $\pi$. As subtle the difference as it may seem, meta-learning-like methods have been proven effective in various applications such as in image restoration \cite{jin2017noise,fan2018decouple,zhang2018dynamically}. See \cite{vanschoren2018meta} for a comprehensive survey on meta-learning. In our experiment section, we will present extensive empirical studies on the generalization ability of the RL-based solver and demonstrate its advantage over some supervised learning based methods.


\item Another purpose of this paper is to raise an awareness of the connection between MDP and numerical PDE solvers, and the general idea of how to use RL to improve PDE solvers or even finding brand new ones. Furthermore, in computational mathematics, a lot of numerical algorithms are sequential, and the computation at each step is expert-designed and usually greedy, e.g., the conjugate gradient method, the fast sweeping method \citep{zhao2005fast}, matching pursuit \citep{mallat1993matching}, etc. We hope our work could motivate more researches in combining RL and computational mathematics, and stimulate more exploration on using RL as a tool to tackle the bottleneck problems in computational mathematics.
\end{itemize}



Our paper is organized as follows. In section \ref{sec::preliminary} we briefly review 1-dimensional conservation laws and the WENO schemes. In section \ref{sec::methods}, we discuss how to formulate the process of numerically solving conservation laws into a Markov Decision Process. Then, we present details on how to train a policy network to mimic human expert in choosing discrete schemes in a spatial-temporary adaptive manner by learning upon WENO. 
In section \ref{sec::experiments}, we conduct numerical experiments on 1-D conservation laws to demonstrate the performance of our trained policy network. Our experimental results show that the trained policy network indeed learned to adaptively choose good discrete schemes that offer better results than the state-of-the-art WENO scheme which is 5th order accurate in space and 4th order accurate in time. This serves as an evidence that the proposed RL framework has the potential to design high-performance numerical schemes for conservation laws in a data-driven fashion. Furthermore, the learned policy network generalizes well to other situations such as different initial conditions, forcing terms, mesh sizes, temporal discrete schemes, etc. To further demonstrate the advantage of using RL to design numerical solvers, we we provide additional comparisons between the WENO learned using RL and two supervised learning based methods, L3D~\cite{bar2019learning} and PINN~\cite{raissi2017physics,raissi2019physics}. The paper ends with a conclusion in section \ref{sec::conclusion}, where possible future research directions are also discussed.

%% file: 02_preliminaries.tex
\section{Preliminaries}
\label{sec::preliminary}
\subsection{Notations}
In this paper, we consider solving the following 1-D conservation laws:
\begin{equation}
\begin{split}
&u_t(x, t) + \left(f(u(x, t))\right)_x = 0 \\
&a \leq x \leq b,  ~t\in[0,T],~~u(x, 0) = u_0(x)
\end{split}
\label{eq::conservationlaw}
\end{equation}
For example, $f= \frac{u^2}{2}$ is the famous Burgers Equation.
We discretize the $(x,t)$-plane by choosing a mesh with spatial size $\Delta x$ and temporal step size $\Delta t$, and define the discrete mesh points $(x_j, t_n)$ by 
\begin{equation}
       \begin{split}
       &x_j = a + j\Delta x,\ t_n = n\Delta t  \\
       &\mbox{with } j = 0, 1, ..., J = \frac{b-a}{\Delta x},\ n = 0, 1, ...,  N=\frac{T}{\Delta t}
       \end{split}
\end{equation}

We denote 
$
    x_{j+\frac12} = x_j + \Delta x/2 = a+(j + \frac12) \Delta x.
$
The finite difference methods will produce approximations $U^n_j$ to 
the solution $u(x_j, t_n)$ on the given discrete mesh points. We denote point-wise values of the true solution to be $u^n_j = u(x_j, t_n)$, and the true point-wise flux values to be $f^n_j = f(u(x_j, t_n))$.

\subsection{WENO -- Weighted Essentially Non-Oscillatory Schemes}
\label{subsec:weno}
WENO (Weighted Essentially Non-Oscillatory) \citep{liu1994weighted} is a family of high order accurate finite difference schemes for solving hyperbolic conservation laws, and has been successful for many practical problems. The key idea of WENO is a nonlinear adaptive procedure that automatically chooses the smoothest local stencil to reconstruct the numerical flux. Generally, a finite difference method solves Eq.\ref{eq::conservationlaw} by using a conservative approximation to the spatial derivative of the flux:
\begin{equation}
    \frac{du_j(t)}{dt} = -\frac{1}{\Delta x}\left(\hat{f}_{j + \frac12} - \hat{f}_{j - \frac12}\right),
    \label{eq::conservation scheme}
\end{equation}
where $u_j(t)$ is the numerical approximation to the point value $u(x_j, t)$ and $\hat{f}_{j + \frac12}$ is the numerical flux generated by a \textbf{numerical flux policy} $$\hat{f}_{j + \frac12} = \pi^{f}(u_{j-r}, ..., u_{j+s}),$$ which is manually designed. Note that the term ``numerical flux policy" is a new terminology that we introduce in this paper, which is exactly the policy we shall learn using RL. In WENO, $\pi^{f}$ works as follows. Using the physical flux values $\{f_{j-2}, f_{j-1}, f_{j}\}$, we could obtain a $3^{th}$ order accurate polynomial interpolation $\hat{f}_{j + \frac12}^{-2}$, where the indices $\{j-2, j-1, j\}$ is called a `stencil'. We could also use the stencil $\{j-1, j, j+1\}$, $\{j, j+1, j+2\}$ or $\{j + 1, j + 2, j + 3\}$ to obtain another three interpolants $\hat{f}_{j + \frac12}^{-1}$, $\hat{f}_{j + \frac12}^{0}$ and $\hat{f}_{j + \frac12}^{1}$. The key idea of WENO is to average (with properly designed weights) all these interpolants to obtain the final reconstruction: 
$$
\hat{f}_{j+\frac12} = \sum_{r = -2}^{1} w_r \hat{f}_{j+1/2}^r, ~~\sum_{r=-2}^{1} w_r = 1.
$$

The weight $w_i$ depends on the smoothness of the stencil. A general principal is: the smoother is the stencil, the more accurate is the interpolant and hence the larger is the weight. To ensure convergence, we need the numerical scheme to be consistent and stable \citep{leveque1992numerical}. It is known that WENO schemes as described above are consistent. For stability, upwinding is required in constructing the flux. The most easy way is to use the sign of the Roe speed $$\bar{a}_{j + \frac12} = \frac{f_{j+\frac12} - f_{j - \frac12}}{ u_{j + \frac12} - u_{j - \frac12}}$$ to determine the upwind direction: if $\bar{a}_{j + \frac12} \geq 0$, we only average among the three interpolants $\hat{f}_{j+\frac12}^{-2}$, $\hat{f}_{j+\frac12}^{-1}$ and $\hat{f}_{j+\frac12}^{0}$; if $\bar{a}_{j + \frac12} < 0$, we use $\hat{f}_{j+\frac12}^{-1}$, $\hat{f}_{j+\frac12}^{0}$ and $\hat{f}_{j+\frac12}^{1}$. 

\paragraph{Some further thoughts.} WENO achieves optimal order of accuracy (up to 5) at the smooth region of the solutions \citep{shu1998essentially}, while lower order of accuracy at singularities. The key of the WENO method lies in how to compute the weight vector $(w_1, w_2, w_3, w_4)$, which primarily depends on the smoothness of the solution at local stencils. In WENO, such smoothness is characterized by handcrafted formula, and was proven to be successful in many practical problems when coupled with high-order temporal discretization. However, it remains unknown whether there are better ways to combine the stencils so that optimal order of accuracy in smooth regions can be reserved while, at the same time, higher accuracy can be achieved near singularities. 
Furthermore, estimating the upwind directions is another key component of WENO, which can get quite complicated in high-dimensional situations and requires lots of logical judgments (i.e. ``if/else"). Can we ease the (some time painful) coding and improve the estimation at the aid of machine learning? 

%% file: 03_algos.tex
\section{Methods}
\label{sec::methods}
In this section we present how to employ reinforcement learning to solve the conservation laws given by \eqref{eq::conservationlaw}. To better illustrate our idea, we first show in general how to formulate the process of numerically solving a conservation law into an MDP. We then discuss how to incorporate a policy network with the WENO scheme. Our policy network targets at the following two key aspects of WENO: \textbf{(1) Can we learn to choose better weights to combine the constructed fluxes? (2) Can we learn to automatically judge the upwind direction, without complicated logical judgments? }
\subsection{MDP Formulation}
\begin{algorithm}
\small
	\SetKwProg{Fn}{Function}{:}{}
    \SetKwData{Input}{Input}{}
    \SetKwData{Output}{Output}{}
    \DontPrintSemicolon
    \Input : initial values $u^0_0, u^0_1, ... , u^0_J$, flux $f(u)$, $\Delta x$, $\Delta t$,  evolve time $N$, left shift $r$ and right shift $s$. \\
    \Output : $\{U^n_j | ~j = 0, ..., J,~  n = 1, ..., N\}$ \\
    $U^0_j = u^0_j, ~j = 0, ..., J$ \\ 
    \For{$n = 1$ \KwTo $N$}
    {
      \For{$j = 0$ \KwTo $J$}
      {
        Compute the numerical flux $\hat{f}^n_{j-\frac12} = \pi^{f}(U^{n-1}_{j-r-1}$, $U^{n-1}_{j-r}$, ..., $U^{n-1}_{j+s-1})$ and $\hat{f}^n_{j + \frac12} = \pi^{f}(U^{n-1}_{j-r}$, $U^{n-1}_{j-r+1}$, ..., $U^{n-1}_{j+s})$, e.g., using the WENO scheme \\
        Compute $\frac{du_j(t)}{dt} 
      = -\frac{1}{\Delta x}(\hat{f}^n_{j + \frac12} - \hat{f}^n_{j - \frac12})$ \\
      Compute $U^{n}_j$ = $\pi^{t}(U^{n-1}_j, \frac{du_j(t)}{dt})$, e.g., using the Euler scheme $U^{n}_j =  U^{n-1}_j + \Delta t \frac{du_j(t)}{dt}$
      }
    }	
  \textbf{Return} $\{U^n_j | ~j = 0, ..., J,~  n = 1, ..., N\}$
    \caption{A Conservation Law Solving Procedure}
    \label{algo::generalprocedure}
\end{algorithm}

As shown in Algorithm \ref{algo::generalprocedure}, the procedure of numerically solving a conservation law is naturally a sequential decision making problem.
The key of the procedure is the numerical flux policy $\pi^f$ and the temporal scheme $\pi^t$ as shown in line 6 and 8 in Algorithm \ref{algo::generalprocedure}. Both policies could be learned using RL. However, in this paper, we mainly focus on using RL to learn the numerical flux policy $\pi^{f}$, while leaving the temporal scheme $\pi^t$ with traditional numerical schemes such as the Euler scheme or the Runge–Kutta methods. A quick review of RL is given in the appendix.

Now, we show how to formulate the above procedure as an MDP and the construction of the state $S$, action $A$, reward $r$ and transition dynamics $P$. Algorithm \ref{algo::RLprocedure} shows in general how RL is incorporated into the procedure. In Algorithm \ref{algo::RLprocedure}, we use a single RL agent. Specifically, when computing $U^n_j$:
\begin{itemize}[leftmargin=*]
    \item The \textbf{state} for the RL agent is $$s^n_j = g_s(U^{n-1}_{j-r-1}, ..., U^{n-1}_{j+s}),$$ where $g_s$ is the state function. The choice of $g_s$ depends on the problem and the choice we made for WENO will be given in Section \ref{sec::experiments}. 
    \item In general, the \textbf{action} of the agent is used to determine how the numerical fluxes $\hat{f}^n_{j + \frac12}$ and $\hat{f}^n_{j - \frac12}$ is computed. In the next subsection, we detail how we incorporate $a^n_j$ to be the linear weights of the fluxes computed using different stencils in the WENO scheme.  
    \item The \textbf{reward} should encourage the agent to generate a scheme that minimizes the error between its approximated value and the true value. Therefore, we define the reward function as $$r^n_j = g_r(U^n_{j-r-1}-u^n_{j-r-1},\cdots,U^n_{j+s}-u^n_{j+s}),$$ e.g., a simplest choice is $g_r = -||\cdot||_2$. Same as $g_s$, the choice of $g_r$ is problem dependent, and the choice we made for WENO will be given in Section \ref{sec::experiments}. 
    \item The \textbf{transition dynamics} $P$ is fully deterministic, and depends on the choice of the temporal scheme at line 10 in Algorithm \ref{algo::RLprocedure}. Note that the next state can only be constructed when we have obtained all the point values in the next time step, i.e., $s^{n+1}_j = g_s(U^{n}_{j-r-1}, ..., U^{n}_{j+s})$ does not only depends on action $a^n_j$, but also on actions $a^n_{j-r-1}, ..., a^n_{j+s}$ (action $a^n_j$ can only determine the value $U^n_j$). This subtlety can be resolved by viewing the process under the framework of multi-agent RL, in which at each mesh point $j$ we use a distinct agent $A^{RL}_j$, and the next state $s^{n+1}_j = g_s(U^{n}_{j-r-1}, ..., U^{n}_{j+s})$ depends on these agents' joint action $\mathbf{a^n_j} = (a^n_{j-r-1}, ..., a^n_{j+s})$. However, it is impractical to train $J$ different agents as $J$ is usually very large, therefore we enforce the agents at different mesh point $j$ to share the same weight, which reduces to case of using just a single agent. The single agent can be viewed as a counterpart of a human designer who decides on the choice of  a \textit{local scheme} based on the current state in traditional numerical methods.
\end{itemize}

\begin{algorithm}
\small
	\SetKwProg{Fn}{Function}{:}{}
    \SetKwData{Input}{Input}{}
    \SetKwData{Output}{Output}{}
    \DontPrintSemicolon
    \Input : initial values $u^0_0, ... , u^0_J$, flux $f(u)$, $\Delta x$, $\Delta t$, evolve time $N$, left shift $r$, right shift $s$ and RL policy $\pi^{RL}$ \\
    \Output : $\{U^n_j | ~j = 0, ..., J,~  n = 1, ..., N\}$ \\
    $U^0_j = u^0_j, ~j = 0, ..., J$ \\ 
    \For{Many iterations}
    {
    Construct initial states $s^0_j = g_s(U^0_{j-r-1}, ..., U^0_{j+s})$ for $j = 0, ..., J$ \\
    \For{$n = 1$ \KwTo $N$}
    {
      \For{$j = 0$ \KwTo $J$}
      {
        Compute the action $a^n_j = \pi^{RL}(s^n_j)$ that determines how $\hat{f}^n_{j + \frac12}$ and $\hat{f}^n_{j - \frac12}$ is computed   \\ 
        Compute $\frac{du_j(t)}{dt} 
       = -\frac{1}{\Delta x}(\hat{f}^n_{j + \frac12} - \hat{f}^n_{j - \frac12})$ \\
        Compute $U^{n}_j$ = $\pi^{t}(U^{n-1}_j, \frac{du_j(t)}{dt})$, e.g., the Euler scheme $U^{n}_j =  U^{n-1}_j + \Delta t \frac{du_j(t)}{dt}$ \\
       Compute the reward $r^n_j = g_r(U^n_{j-r-1}-u^n_{j-r-1},\cdots,U^n_{j+s}-u^n_{j+s})$.
      }
      
      Construct the next states $s^{n+1}_j = g_s(u^n_{j-r-1}, ..., u^n_{j+s})$ for $j = 0, ... ,J$ \\
      Use any RL algorithm to train the RL policy $\pi^{RL}$ with the transitions $\{(s^n_j, a^n_j, r^n_j, s^{n+1}_j)\}_{j=0}^J$. \\
    }	
    }
  \textbf{Return} the well-trained RL policy $\pi^{RL}$.
    \caption{General RL Running Procedure}
    \label{algo::RLprocedure}
\end{algorithm}

\subsection{Improving WENO with RL}\label{RL-WENO}
We now present how to transfer the actions of the RL policy to the weights of WENO fluxes. Instead of directly using $\pi^{RL}$ to generate the numerical flux, we use it to produce the weights of numerical fluxes computed using different stencils in WENO. Since the weights are part of the configurations of the WENO scheme, our design of action essentially makes the RL policy a meta-learner, and enables more stable learning and better generalization power than directly generating the fluxes.

Specifically, at point $x_j$ (here we drop the time superscript $n$ for simplicity), to compute the numerical flux $\hat{f}_{j - \frac12}$ and $\hat{f}_{j + \frac12}$, we first construct four fluxes $\{\hat{f}^{i}_{j  - \frac12}\}_{i=-2}^1$ and $\{\hat{f}^{i}_{j + \frac12}\}_{i=-2}^1$ using four different stencils just as in WENO, and then use the RL policy $\pi^{RL}$ to generate the weights of these fluxes: $$\pi^{RL}(s_j) = \left(w_{j-\frac12}^{-2}, ..., w_{j-\frac12}^{1}, w_{j+\frac12}^{-2}, ..., w_{j+\frac12}^{1}\right),\quad\mbox{with}\ \sum_{i=-2}^1w_{j\pm\frac12}^{i}=1.$$
The numerical flux is then constructed by averaging these fluxes:
$$\hat{f}_{j - \frac12} = \sum_{i = -2}^1 w_{j-\frac12}^{i}\hat{f}^{i}_{j - \frac12}\quad \mbox{and}\quad \hat{f}_{j + \frac12} = \sum_{i = -2}^1 w_{j + \frac12}^i\hat{f}^{i}_{j + \frac12}.$$

Note that the determination of upwind direction is automatically embedded in the RL policy since it generates four weights at once. For instance, when the roe speed $\bar{a}_{j+\frac12} \geq 0$, we expect the $4^{th}$ weight $w_{j + \frac12}^1\approx 0$ and when $\bar{a}_{j+\frac12} < 0$, we expect $w_{j + \frac12}^{-2}\approx 0$. Note that the upwind direction can be very complicated in a system of equations or in the high-dimensional situations, and using the policy network to automatically embed such a process could save lots of efforts in algorithm design and implementation. Our numerical experiments show that $\pi^{RL}$ can indeed automatically determine upwind directions for 1D scalar cases. Although this does not mean that it works for systems and/or in high-dimensions, it shows the potential of the proposed framework and value for further studies.

%% file: 04_experiments.tex
\section{Experiments}

 \begin{table*}[t]
\small
\centering
\begin{tabular}{|c|c|c|c|c|c|c|}
\hline
\multirow{2}{*}{\diagbox{$\Delta t$}{$\Delta x$}} & \multicolumn{2}{c|}{0.02} & \multicolumn{2}{c|}{0.04} & \multicolumn{2}{c|}{0.05} \\ \cline{2-7} 
   & RL-WENO         & WENO         & RL-WENO         & WENO         & RL-WENO         & WENO         \\ \hline
0.002 & 10.81 (3.96) & 11.13 (3.83) & 18.79 (12.44) & 19.45 (9.32) & 28.61 (33.37) & 35.95 (27.7) \\\hline
0.003 & 10.83 (3.96) & 11.14 (3.82) & 18.82 (12.48) & 19.47 (9.31) & 28.62 (33.34) & 35.96 (27.67) \\\hline
0.004 & 10.83 (3.96) & 11.14 (3.84) & 18.89 (12.69) & 19.48 (9.33) & 28.61 (33.27) & 35.93 (27.63) \\\hline
0.005 & 10.85 (3.96) & 11.15 (3.84) & 18.96 (12.84) & 19.52 (9.35) & 28.48 (33.04) & 35.93 (27.61) \\\hline
0.006 & 10.89 (3.93) & 11.16 (3.83) & 18.95 (12.79) & 19.51 (9.3) & 28.58 (33.08) & 35.89 (27.5) \\\hline
 \end{tabular}
 \caption{Training setting: inviscid. Test setting: inviscid. Comparison of relative errors ($\times 10^{-2}$) of RL-WENO and WENO with the means and standard deviations of the errors among 25 random test initial conditions in the parenthesis. Temporal discretization: RK4; flux function: $\frac12 u^2$. 
 } 
 \label{tab:Linear-RL-compare-rk4-u2}
 \end{table*}

\begin{table*}[t]
\small
\centering
\begin{tabular}{|c|c|c|c|c|c|c|}
\hline
\multirow{2}{*}{\diagbox{$\Delta t$}{$\Delta x$}} & \multicolumn{2}{c|}{0.02} & \multicolumn{2}{c|}{0.04} & \multicolumn{2}{c|}{0.05} \\ \cline{2-7} 
  & RL-WENO         & WENO         & RL-WENO         & WENO         & RL-WENO         & WENO         \\ \hline
0.002 & 10.05 (2.74) & 10.25 (2.65) & 16.89 (3.59) & 17.09 (3.56) & 17.07 (4.2) & 17.4 (4.37) \\\hline
0.003 &- & 10.26 (2.65) & 16.9 (3.59) & 17.1 (3.56) & 17.09 (4.2) & 17.42 (4.37) \\\hline
0.004 &- &- & 16.9 (3.6) & 17.1 (3.56) & 17.1 (4.2) & 17.43 (4.37) \\\hline
0.005 &- &- &- & 17.12 (3.56) & 17.11 (4.22) & 17.43 (4.38) \\\hline
\end{tabular}

\caption{Training setting: inviscid. Test setting: inviscid. Comparison of relative errors ($\times 10^{-2}$) of RL-WENO and WENO with the means and standard deviations of the errors among 25 random test initial conditions. Temporal discretization: RK4; flux function: $\frac{1}{16} u^4$. 
} 
\label{tab:Linear-RL-compare-rk4-u4}
\end{table*}

\label{sec::experiments}
In this section, we describe the training and testing of the proposed RL conservation law solver, and compare it with WENO and some recently proposed supervise learning-based methods for solving conservation laws: L3D~\cite{bar2019learning} and PINNs~\cite{ raissi2017physics,raissi2019physics}.


\subsection{Training and Testing Setup}
\label{sec:weno_exp_setup}
In this subsection, we explain the general training and testing setup. We train the RL policy network on the Burgers equation, whose flux is computed as $f(u) = \frac12u^2$. In all the experiments, we set the left-shift $r = 2$ and the right shift $s = 3$. The state function $g_s(s_j) = g_s(U_{j-r-1}, ..., U_{j+s})$ will generate two vectors: $$s^l = (f_{j-r-1}, ..., f_{j+s-1}, \bar{a}_{j - \frac12})\quad \mbox{and}\quad s^r = (f_{j-r}, ..., f_{j+s}, \bar{a}_{j + \frac12})$$ for computing $\hat{f}_{j - \frac12}$ and $\hat{f}_{j + \frac12}$ respectively. $s_l$ and $s_r$ will be passed into the same policy neural network $\pi^{RL}_\theta$ to produce the desired actions, as described in Section \ref{RL-WENO}. The reward function $g_r$ simply computes the infinity norm, i.e., $$g_r(U_{j-r-1}-u_{j-r-1}, ..., U_{j+s}-u_{j+s}) = -||(U_{j-r-1}-u_{j-r-1}, ..., U_{j+s}-u_{j+s})||_{\infty}.$$ 
The policy network $\pi^{RL}_\theta$ is a feed-forward Multi-layer Perceptron with $6$ hidden layers, each has 64 neurons and use ReLU \citep{goodfellow2016deep} as the activation function. We use the Twin Delayed Deep Deterministic (TD3) policy gradient algorithm \citep{fujimoto2018addressing} to train the RL policy. 

We consider the following three settings of the Burgers Equation:
\begin{equation}
    u(x, t)_t + (\frac12 u(x,t)^2)_x = \eta u(x, t)_{xx} + F(x, t), ~~~~u(x, 0) = u_0(x), ~~x \in D, ~~t \in [0, T]   
\end{equation}

\begin{itemize}
    \item \textbf{Inviscid}: inviscid Burgers Equation with different random initial conditions:
    \begin{equation}
        \begin{split}
                & ~~~~~~~~~~~~~~~~~~~~~ \eta = 0, ~~F(x, t) = 0 \\
                & ~~~~~~~~~~ u_0(x) = a + b \cdot \sin(c \pi x) + d \cdot \cos(e \pi x) \\
& |a| + |b| + |d| = 4, ~|a| \leq 1.2, ~|b| \leq 3 - |a|,  ~c \in \{4, 6, 8\}.
        \end{split}
    \label{eq:initial_condition}
    \end{equation}
    The reason for designing such kind of initial conditions is to ensure they have similar degree of smoothness and thus similar level of difficulty in learning. For training, we randomly generate $25$ such initial conditions, and we always set $c = 8$ to ensure enough shocks in the solutions. For testing, we randomly generate another $25$ initial conditions in the same form with $c$ chosen in $\{4, 6\}$. During training, the computation domain is $D = [-1, 1]$, and the terminal time $T$ is randomly sampled from $[0.25,1]$ with equal probability; during test, the terminal time is fixed at $T = 0.9$.  We train on two different grid sizes : $(\Delta x, \Delta t) \in \{( 0.02, 0.002), (0.04, 0.004)\}$.
        \item \textbf{Forcing}\footnote{This setting is exactly the same as the training/testing setting used in L3D~\cite{bar2019learning}.}: Viscous Burgers Equation with forcing, and a fixed initial value.
      \begin{equation}
        \begin{split}
            & ~~~~~~~~~~~~~~~~~~~\eta \in \{0.01, 0.02, 0.04\}, ~~u_0(x) = 0 \\
            & ~~~~~~~~~~~~F(x, t) = \sum_{i=1}^N A_i \sin(\omega_i t + 2 \pi l_i x / L + \psi_i) \\
            & N = 20, ~A_i \in[-0.5, 0.5], ~\omega_i \in [-0.4, 0.4], ~\phi_i \in [0, 2 \pi], ~l_i \in \{3, 4, 5, 6\}
        \end{split}
    \label{eq:l3d_condition}
\end{equation}
        During training, $\eta$ is fixed at $0.01$, and we test with $\eta$ in $\{0.01, 0.02, 0.04\}$. 50 random forcings are generated, the first half is used for training, and the second half is used for testing. During training, the computation domain is $D = [0, 2\pi]$, and the terminal time $T$ is randomly sampled from $[0.25\pi,\pi]$ with equal probability. The test terminal time is fixed at $T = 0.9\pi$. We train on two different grid sizes: $(\Delta x, \Delta t) \in \{( 0.02\pi, 0.002\pi), (0.04\pi, 0.004\pi)\}$.

    \item \textbf{Viscous}\footnote{Note that the viscous Burgers Equation is easier to solve compared with the inviscid version, as the viscous term smooths out the solution and helps remove shocks in the solution.}: Viscous Burgers Equation with different random initial conditions: $\eta \in \{0.01, 0.02, 0.04\}$, and $F(x,t) = 0$.  The viscous term is handled using central difference schemes. The initial conditions are generated in the same form as in equation~\ref{eq:initial_condition}. For both training and testing, we sample 25 initial conditions with $c$ in $\{4, 6\}$. For training, we fix $\eta = 0.01$, while for testing, we test with different values of $\eta$ in $\{0.01, 0.02, 0.04\}$.  The computation domain, and training grid sizes are the same as the inviscid version.

\end{itemize}

When training the RL agent, we use the Euler scheme for temporal discretization. During test, we always use the 4th order Runge-Kutta (RK4) for temporal discretization. The true solution needed for computing the reward and for evaluation is generated using WENO on the same computation domain with a much finer grid ($\Delta x = 0.002$, $\Delta t = 0.0002$) and use the RK4 as the temporal discretization. 
In the following, we denote the policy network that generates the weights of the WENO fluxes (as described in Section \ref{RL-WENO}) as RL-WENO. In all the settings, we compare RL-WENO with WENO. As L3D is (only) demonstrated to work in the Forcing setting, we also compare RL-WENO with L3D in this setting. 
We also compare RL-WENO with PINNs in both the inviscid and viscous setting. 
To demonstrate the ability to perform out-of-distribution generalization of RL-WENO,
we train RL-WENO on one of the three settings and test it on the other two settings without fine-tuning or re-training.
For all experiments, the evaluation metric is the relative L2 error compared with the true solution computed using WENO on a finer grid as described above. See Table~\ref{tab:experiment_summary} for a summary and road-map for all the experiments.

We also give a summary of the inference time for RL-WENO, WENO, L3D, and PINNs in Table~\ref{tab:computation_time}. RL-WENO, WENO and L3D are tested in the forcing setting, averaged over 25 random forcings, and PINNs is tested in the viscous setting, averaged over 25 different random initial conditions. For PINNs, as it needs to be trained on each different initial condition, the inference time is the same as the training time.

\begin{table}[t]
\centering
\small
\begin{tabular}{|c|c|c|c|}
\hline
\diagbox{Train}{Test} &
\begin{tabular}[c]{@{}c@{}} inviscid w/o forcing \\ random initial value \end{tabular}
   &
\begin{tabular}[c]{@{}c@{}}viscous w/ forcing 
\\ fixed initial value\end{tabular}
   &
   \begin{tabular}[c]{@{}c@{}}viscous w/o forcing \\ random initial value\end{tabular}
  \\ \hline
\begin{tabular}[c]{@{}c@{}}  inviscid w/o forcing \\ random initial value \end{tabular} &

\begin{tabular}[c]{@{}c@{}}RL-WENO \cmark ( Table~\ref{tab:Linear-RL-compare-rk4-u2}, \ref{tab:Linear-RL-compare-rk4-u4}) \\ L3D ? \\ PINNs \semichecked (Table~\ref{tab:compare_PINNs}) \end{tabular} &
  \begin{tabular}[c]{@{}c@{}}RL-WENO \cmark ( Table~\ref{tab:generalize_to_forcing}) \\ L3D ? \end{tabular} &
  \begin{tabular}[c]{@{}c@{}}RL-WENO \cmark ( Table~\ref{tab:generalize_to_viscous})\\ L3D ? \\ \end{tabular} \\ \hline 
 
\begin{tabular}[c]{@{}c@{}}viscous w/ forcing 
\\ fixed initial value\end{tabular}  &
  \begin{tabular}[c]{@{}c@{}}RL-WENO \cmark (Table~\ref{tab:forcing-rl-weno-on-viscous-and-inviscid})  \\ L3D \xmark \end{tabular} &
  \begin{tabular}[c]{@{}c@{}}RL-WENO \cmark (Table~\ref{tab:compare_l3d}) \\ L3D \cmark (Table~\ref{tab:compare_l3d}) 
  \end{tabular} &
  \begin{tabular}[c]{@{}c@{}}RL-WENO \cmark (Table~\ref{tab:forcing-rl-weno-on-viscous-and-inviscid}) \\ L3D \xmark \end{tabular}\\ \hline
  
\begin{tabular}[c]{@{}c@{}}viscous w/o forcing \\ random initial value\end{tabular} &
  \begin{tabular}[c]{@{}c@{}}RL-WENO  \cmark (Table~\ref{tab:viscous-rl-weno-on-viscous-and-inviscid}) \\ L3D ? \end{tabular} &
  \begin{tabular}[c]{@{}c@{}}RL-WENO \cmark (Table~\ref{tab:viscous-rl-weno-on-forcing}) \\ L3D ? \end{tabular} &
  \begin{tabular}[c]{@{}c@{}}RL-WENO \cmark (Table~\ref{tab:viscous-rl-weno-on-viscous-and-inviscid}) \\ L3D ? \\ PINNs \semichecked (Table~\ref{tab:compare_PINNs}) \end{tabular}\\ \hline

\end{tabular}
\caption{A summary for all the experiments we conducted. \cmark~means the method performs well in that case. \semichecked means the method performs relatively well.  \xmark~means it fails. ? means that we failed to train a decent model in this setting.  }\label{tab:experiment_summary}
\end{table}

\begin{table}[]
    \centering
    \begin{tabular}{|c|c|c|c|c|}
    \hline
    $(\Delta x, \Delta t)$     & RL-WENO & WENO & L3D & PINN \\\hline
    $(0.02, 0.002)$     & 2.23 & 4.14 & 0.59 & 2148.34 \\\hline
    $(0.04, 0.004)$ & 0.87 & 1.09  & 0.51 & 2024.14\\\hline
    $(0.05, 0.005)$ & 0.6 & 0.71 & 0.5 &  2058.66\\ \hline
    \end{tabular}
    \caption{Inference time (in seconds) of different methods. The terminal time is $T = 0.9$. }
    \label{tab:computation_time}
\end{table}

\begin{figure*}[t]
    \begin{center}
    \begin{tabular}{cc}
   \includegraphics[width=0.45\textwidth]{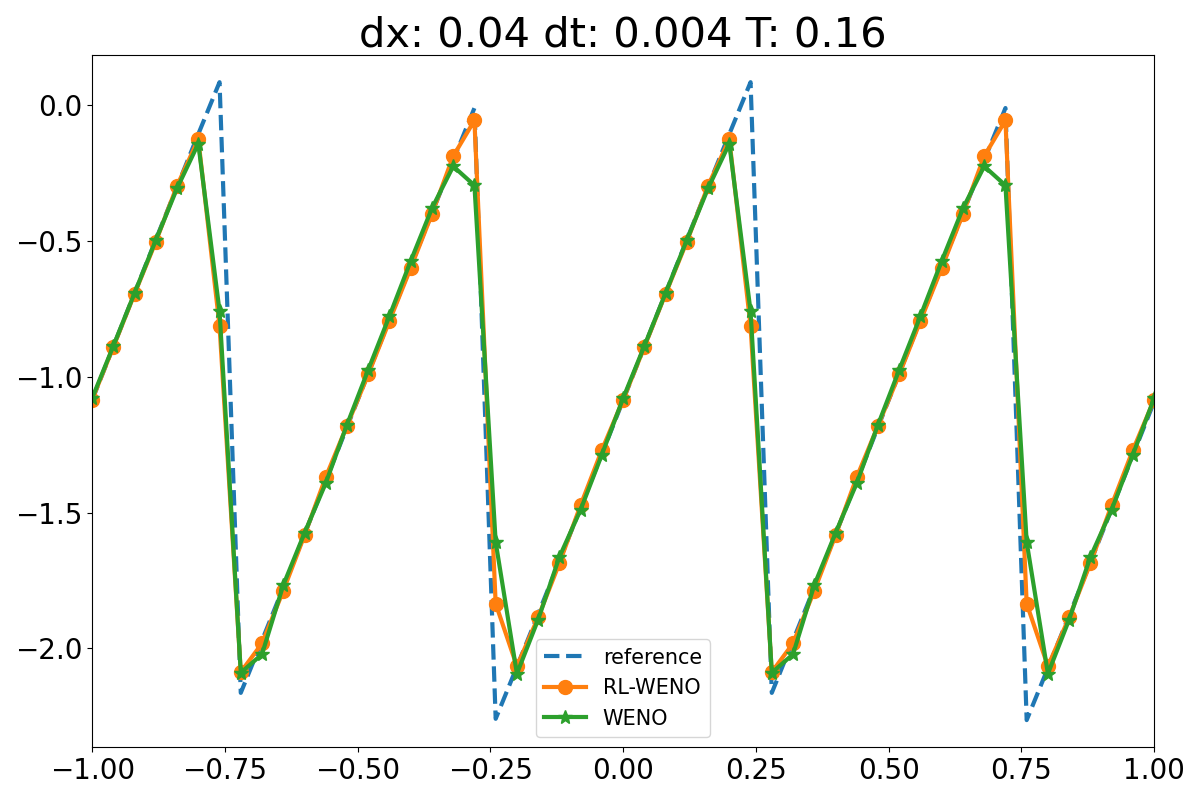} & 
   \includegraphics[width=0.45\textwidth]{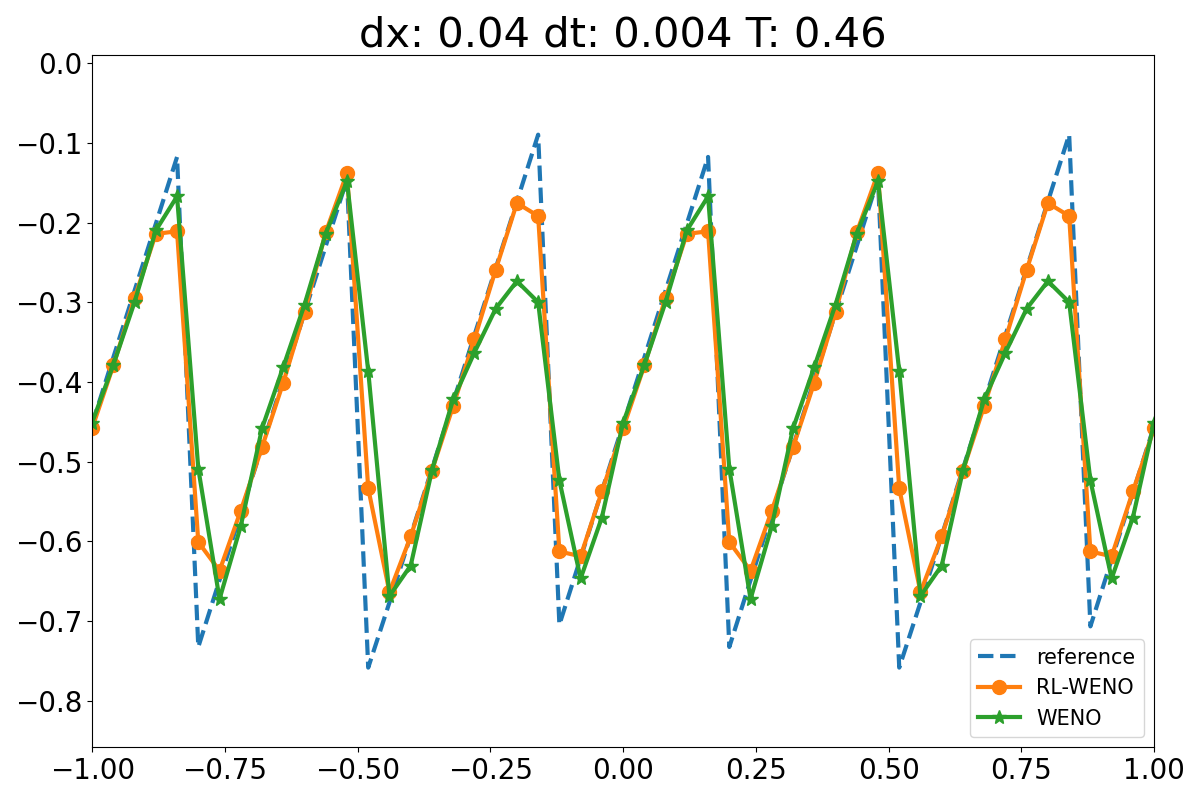}
    \\
    \includegraphics[width=0.45\textwidth]{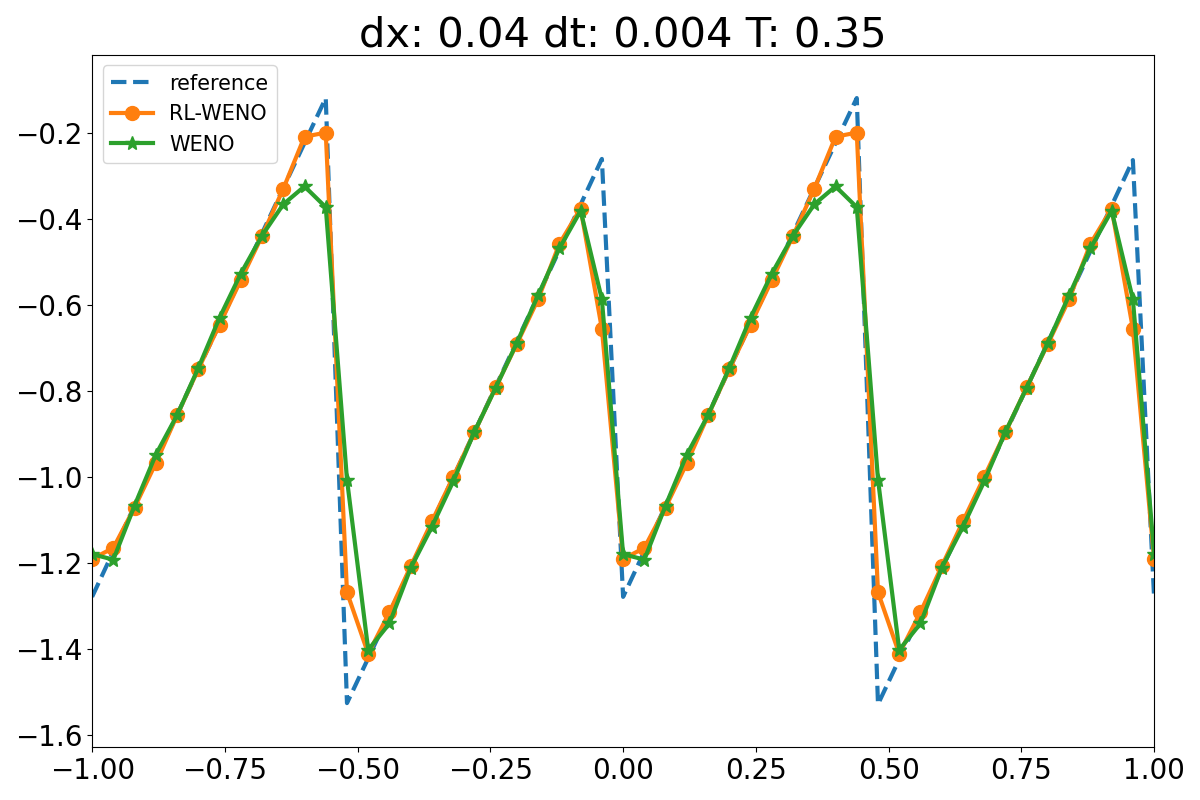} &
        \includegraphics[width=0.45\textwidth]{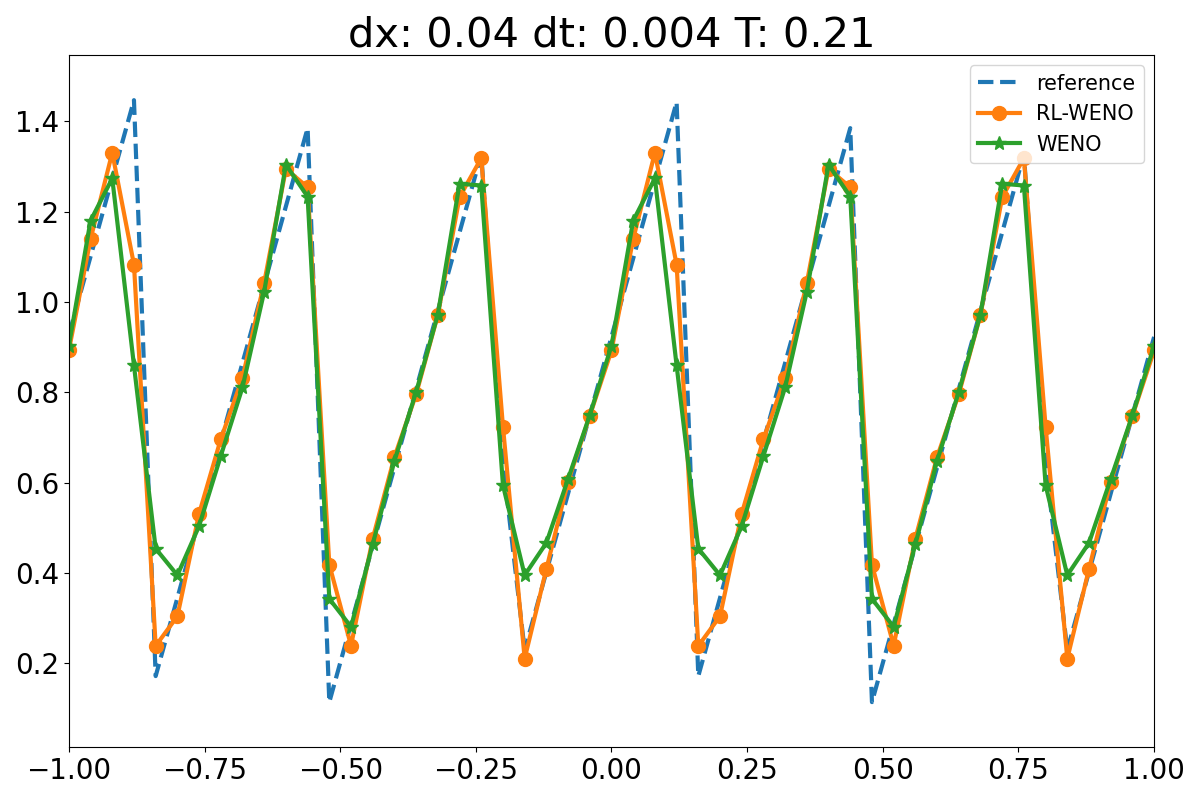} \\
    \end{tabular}
	\caption{Solutions of RL-WENO (orange), WENO (green) and exact solutions (dashed blue) on 4 different initial conditions. RL-WENO demonstrates clear advantages at singularities.}
	\label{fig:Linear-RL-compare}
    \end{center}
\end{figure*}

\begin{figure*}[t]
    \begin{center}
    \includegraphics[width=\textwidth]{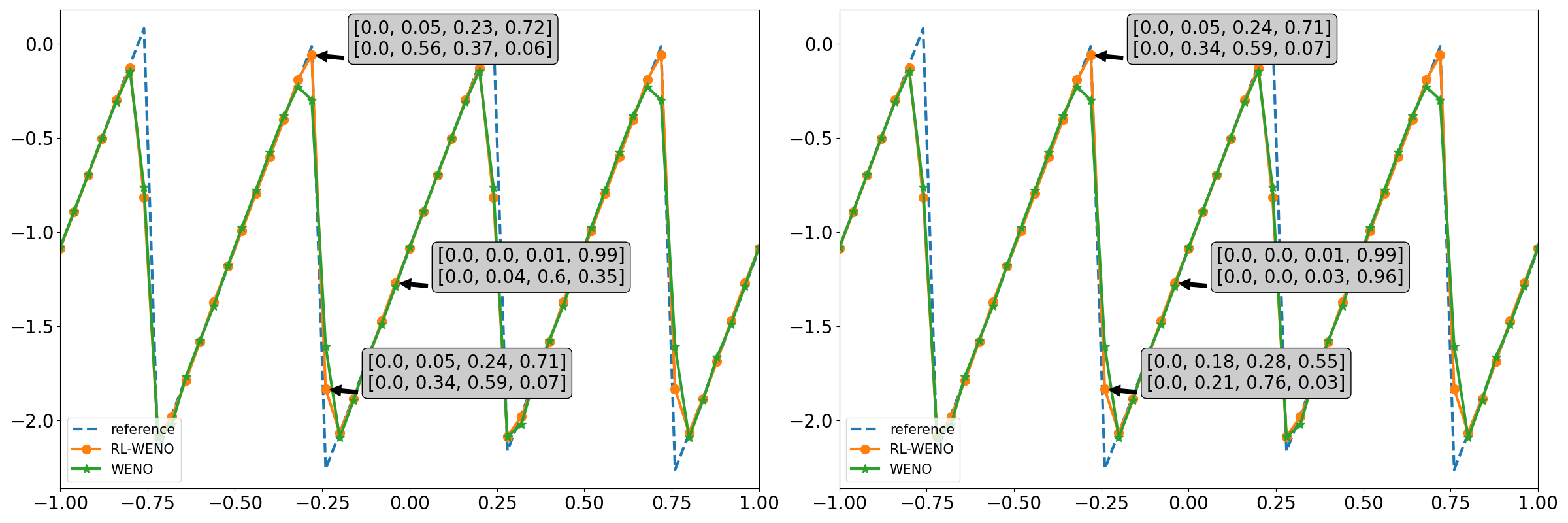}
	\caption{This figure compares the weights generated by RL-WENO $\pi^{RL}$ and those of WENO. The weights shown in the left plot are $\{w_{j-\frac12}^r\}_{r=-2}^1$; while those in the right plot are $\{w_{j+\frac12}^r\}_{r=-2}^1$. In each of the two plots, the 4 numbers in the upper bracket of each location are the weights of RL-WENO and those in the lower bracket are the weights of WENO. We see that RL-WENO correctly judges the upwind direction, and generates the weight in a different style compared with WENO. }
	\label{fig:weights-compare}
    \end{center}
\end{figure*}

\begin{figure*}[t]
    \centering
    \begin{tabular}{cc}
        

    \includegraphics[width=0.4\textwidth]{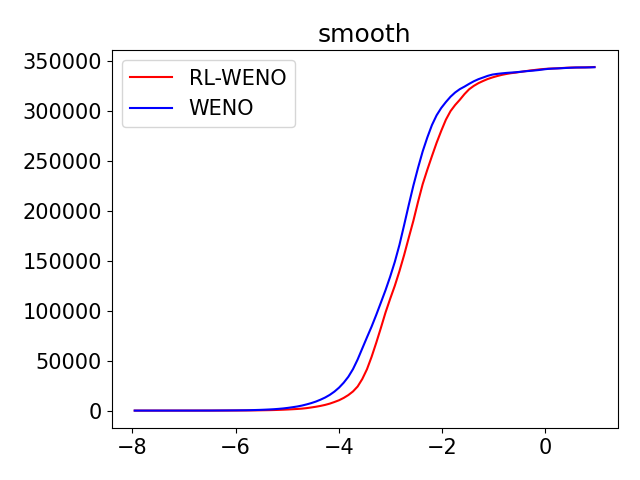} &
    \includegraphics[width=0.4\textwidth]{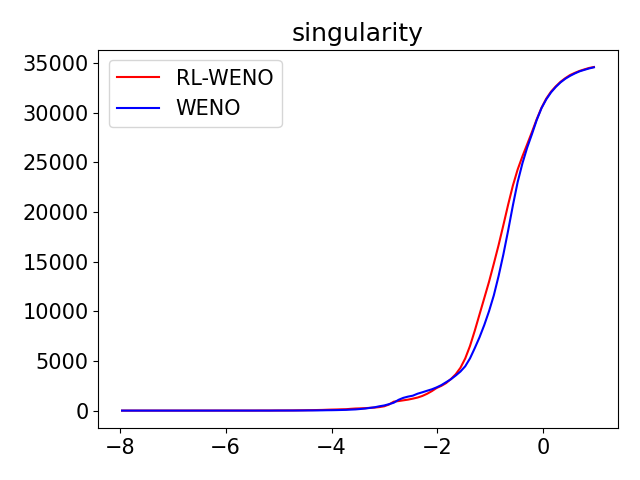} \\
        
    (a) Smooth regions, $f(u) = \frac{1}{2}u^2$  & (b) Near singularities, $f(u) = \frac{1}{2}u^2$  \\

    \end{tabular}
	\caption{These figures show the total number of grid points whose error is under a specific value (i.e. the accumulated distribution function). The $x$-axis is the error in logarithmic (base 10) scale. (a) in smooth regions, (b) near singularities. }
	\label{fig:smooth-singular1}
\end{figure*}

\begin{figure*}[t]
    \centering
    \includegraphics[width=\textwidth]{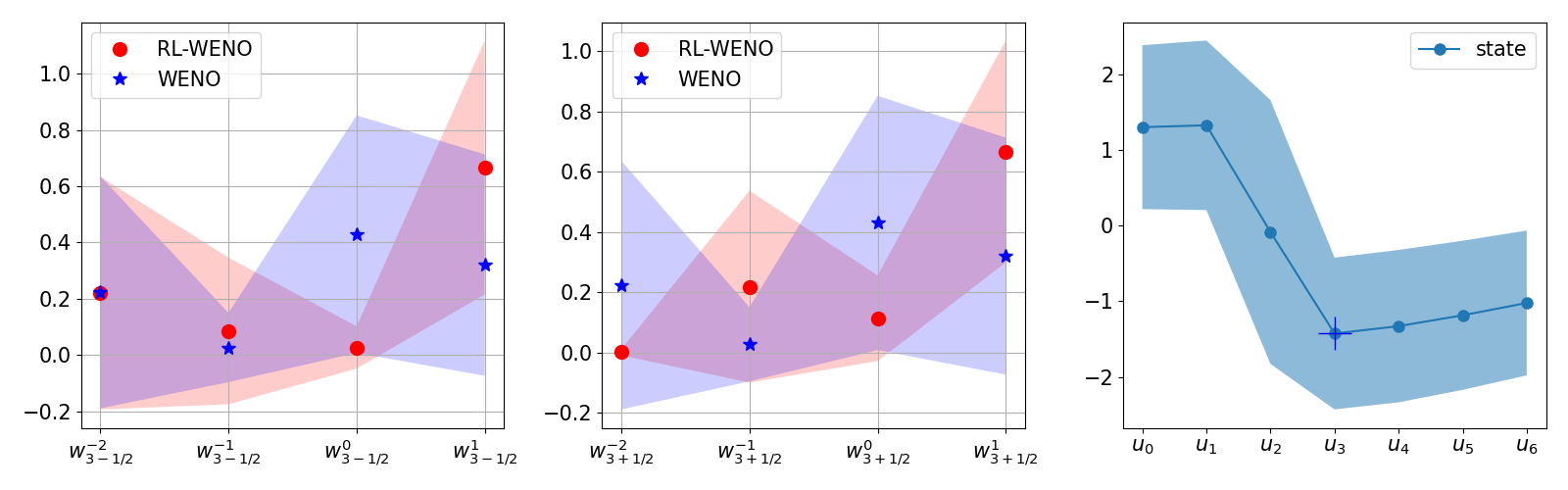}
    \caption{The distribution of weights computed by RL-WENO and WENO near singularities. The singular input states are shown in the third plot, while the first two plots show the weights at $u_3$: $\{w_{3\pm\frac12}^{j}\}_{j=1}^4$. In total 12215 input states are generated, and the shaded region is the standard deviation. }
    \label{fig:singular_distribution}
\end{figure*}

\subsection{inviscid Burgers Equation with random initial conditions}

\para{Generalization to new mesh sizes, flux function, and temporal discretization.} We first test whether the RL policy trained in this setting can generalize to different temporal discretization schemes, mesh sizes and flux functions that are not included in training. Table \ref{tab:Linear-RL-compare-rk4-u2} and Table  \ref{tab:Linear-RL-compare-rk4-u4} present the comparison results, where the number shows the relative error (computed as $\frac{||U - u||_{2}}{||u||_2}$ with the 2-norm taking over all $x$) between the approximated solution $U$ and the true solution $u$, averaged over all evolving steps ($T = 0.9$) and 25 random test initial values. Numbers in the bracket shows the standard deviation over the 25 initial values. Several entries in the table are marked as `-' because the corresponding CFL number is not small enough to guarantee convergence. 

Our experimental results show that, compared with the high order accurate WENO (5th order accurate in space and 4th order accurate in time), the linear weights learned by RL not only achieves smaller errors, but also generalizes well to: 
\begin{itemize}
    \item New time discretization scheme: trained with Euler, tested with RK4;
    \item New mesh sizes: see Table \ref{tab:Linear-RL-compare-rk4-u2} and Table  \ref{tab:Linear-RL-compare-rk4-u4} for results of varied $\Delta x$ and $\Delta t$. Recall we only train with $(\Delta x, \Delta t)=\{(0.02, 0.002), (0.04, 0.004)\}$;
    \item  New flux function: trained with $f(u)=\frac12 u^2$ (results shown in Table \ref{tab:Linear-RL-compare-rk4-u2}), tested with $\frac1{16}u^4$ (results shown in Table  \ref{tab:Linear-RL-compare-rk4-u4}).
\end{itemize} 

Figure \ref{fig:Linear-RL-compare} shows some examples of the solutions. As one can see, the solutions generated by RL-WENO not only achieve the same accuracy as WENO at smooth regions, but also have clear advantage over WENO near singularities which is particularly challenging for numerical PDE solvers and important in applications.

\para{More analysis on how RL-WENO chooses the weights.} First, as shown in Figure \ref{fig:weights-compare}, RL-WENO can indeed correctly determine upwind directions and generate local numerical schemes in an adaptive fashion. More interestingly, Figure \ref{fig:weights-compare} further shows that comparing to WENO, RL-WENO seems to be able to select stencils in a different way from it, and eventually leads to a more accurate solution. 


As mentioned in Section \ref{subsec:weno}, WENO itself already achieves an optimal order of accuracy in the smooth regions. Since RL-WENO can further improve upon WENO, it should have achieved higher accuracy especially near singularities. Therefore, we provide additional demonstrations on how RL-WENO performs in the smooth/singular regions. We run RL-WENO and WENO on a set of initial conditions, and record the approximation errors at every spatial-temporal location. We then separate the errors in the smooth and singular regions, and compute the distribution of the errors on the entire spatial-temporal grids with multiple initial conditions. The results are shown in figure \ref{fig:smooth-singular1}. In figure \ref{fig:smooth-singular1}, the $x$-axis is the logarithmic (base 10) value of the error and the y-axis is the number of grid points whose error is less than the corresponding value on the $x$-axis, i.e., the accumulated distribution of the errors. The results illustrate that RL-WENO indeed performs better than WENO near singularities. 

Given that RL-WENO performs better than WENO at singularities, we further investigate how it inferences the weights differently from WENO. Figure \ref{fig:singular_distribution} shows the distribution of weights computed by both methods near singularities. Again, it can be seen that RL-WENO generates the weights in a different way from WENO. Interestingly, the smaller standard deviation of RL-WENO also seems to show that RL-WENO is more confident on its choice of the weights. These results suggest that the proposed RL framework has the potential to surpass human experts in designing numerical schemes for conservation laws.

\para{Comparison with PINNs}

\begin{table}[t]
\centering
\begin{tabular}{|c|c|c|c|c|}
\hline
$\eta$                & $\Delta x$ & RL-WENO & WENO & PINNs \\ \hline
\multirow{3}{*}{0} & 0.02       & 10.81 (3.96) & 11.13 (3.83) & 40.06(18.82) \\ \cline{2-5} 
                      & 0.04       & 18.89 (12.69) & 19.48 (9.33) & 51.61(23.68) \\ \cline{2-5} 
                      & 0.05      & 28.48 (33.04) & 35.93 (27.61) & 51.18(18.31)    \\ \hline
\multirow{3}{*}{0.01} & 0.02       & 1.64 (0.57) & 1.94 (0.55) & 34.09(26.92) \\ \cline{2-5} 
                      & 0.04       & 4.29 (0.93) & 4.93 (0.73) & 48.28(29.71) \\ \cline{2-5} 
                      & 0.05      & 7.6 (3.94) & 7.66 (1.71) & 47.95(19.86)   \\ \hline
\multirow{3}{*}{0.02} & 0.02       & 0.74 (0.35) & 0.87 (0.37) & 31.13(34.35)     \\ \cline{2-5} 
                      & 0.04       & 1.91 (0.58) & 2.33 (0.51) & 47.41(35.84)    \\ \cline{2-5} 
                      & 0.05       & 2.97 (1.07) & 3.64 (0.77) & 43.05(20.1)) \\ \hline
\multirow{3}{*}{0.04} & 0.02       & 0.42 (0.21) & 0.41 (0.21) & 34.98(37.83)     \\ \cline{2-5} 
                      & 0.04       & 0.87 (0.38) & 0.96 (0.36) & 52.27(44.74)    \\ \cline{2-5} 
                      & 0.05       & 1.38 (0.61) & 1.52 (0.5) & 47.0(24.47)     \\ \hline
\end{tabular}
\caption{Comparison of RL-WENO, WENO and PINNs in the inviscid and viscous settings with different viscosity and grid sizes. $\Delta t$ is computed as $0.1 \times \Delta x$. The numbers report relative errors ($\times 10^{-2}$) of RL-WENO, WENO and L3D with the means and standard deviations of the errors among 25 random test initial functions. When $\eta = 0$, the RL-WENO policy is trained in the inviscid setting. For $\eta > 0$, the RL-WENO policy is trained in the viscous setting. }
\label{tab:compare_PINNs}
\end{table}

\begin{figure}
    \centering
    \includegraphics[width=0.45\textwidth]{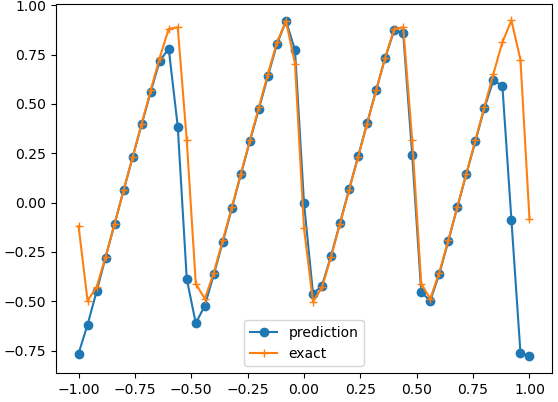}
    \includegraphics[width=0.45\textwidth]{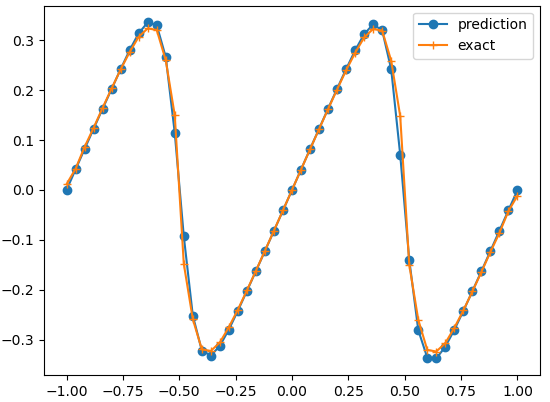}
    \caption{Comparison of PINNs trained on a complex initial condition (left) and a simple initial condition (right). The complex initial condition is $u_0(x) = 1.2 \sin(2 \pi  x) + 0.4 + \cos(2 \pi x)$, while the simple initial condition is $u_0(x) = \sin(2\pi x)$. The viscosity is $\eta = 0.02$. The terminal times are $T=0.2$ (left) and $T=0.8$ (right).}
    \label{fig:PINNs}
\end{figure}

PINNs~\cite{raissi2017physics} directly uses a deep neural network to approximate the solution $u$ to the conservation law \eqref{eq::conservationlaw}. Therefore, it needs to be trained individually for each different initial condition, as the underlying solution is different. The implementation of PINNs is based on the released codes by the authors \footnote{https://github.com/maziarraissi/PINNs}. Note that we assume periodic boundary condition of \eqref{eq::conservationlaw}. Therefore, we modified the released codes of PINNs according to another related online codes \footnote{https://github.com/MinglangYin/PyTorchTutorial/tree/master/PINNs}.
The comparison result is shown in Table~\ref{tab:compare_PINNs}. In fact, PINNs performs poorly with relatively complicated initial conditions such as the ones we chose in Eq.~\eqref{eq:initial_condition}. Indeed, when we simplify the initial condition, e.g., $u_0(x) = \sin(2\pi x)$, the performance of PINNs is significantly improved (see a comparison in Figure~\ref{fig:PINNs}). To resolve this issue, we may choose more expressive neural networks to approximate the solutions. However, this may also increase the difficulty of training. 

\para{Out-of-distribution generalization to the other two test settings. } We now test if RL-WENO trained in the inviscid setting can generalize to the other two settings: viscous and forcing.  Table~\ref{tab:generalize_to_forcing} and~\ref{tab:generalize_to_viscous} show the results.  Surprisingly, RL-WENO generalizes relatively well to these two test settings without any re-training or fine-tuning. We attribute the good generalization ability of RL-WENO to our careful action design, which essentially makes RL-WENO a meta-learner under the WENO framework and thus have strong out-of-distribution generalization.

\begin{table}[t]
\footnotesize
\centering
\begin{tabular}{|c|c|c|c|c|c|c|}
\hline
\multirow{2}{*}{\diagbox{$\Delta x$}{$\eta$}} & \multicolumn{2}{c|}{0.01} & \multicolumn{2}{c|}{0.02} & \multicolumn{2}{c|}{0.04} \\ \cline{2-7} 
   & RL-WENO         & WENO         & RL-WENO         & WENO         & RL-WENO         & WENO         \\ \hline
0.02 & 4.75 (0.73) & 4.74 (0.7) & 2.38 (0.46) & 2.59 (0.45) & 1.07 (0.2) & 1.02 (0.27) \\\hline
0.04 & 10.78 (1.4) & 10.3 (1.35) & 6.95 (1.15) & 6.6 (0.95) & 3.84 (0.7) & 3.68 (0.62) \\\hline
0.05 & 13.97 (1.93) & 13.55 (2.07) & 9.76 (1.44) & 9.33 (1.4) & 5.67 (0.89) & 5.42 (0.81) \\\hline
 \end{tabular}
\caption{Training setting: inviscid. Test setting: forcing. The numbers report relative errors ($\times 10^{-2}$) of RL-WENO and WENO with the means and standard deviations of the errors among 25 random test forcings. We see that RL-WENO can generalize relatively well to this new test setting and achieves similar performance to WENO.}
\label{tab:generalize_to_forcing}
\end{table}

\begin{table}[t]
\footnotesize
\centering
\begin{tabular}{|c|c|c|c|c|c|c|}
\hline
\multirow{2}{*}{\diagbox{$\Delta x$}{$\eta$}} & \multicolumn{2}{c|}{0.01} & \multicolumn{2}{c|}{0.02} & \multicolumn{2}{c|}{0.04} \\ \cline{2-7} 
   & RL-WENO         & WENO         & RL-WENO         & WENO         & RL-WENO         & WENO         \\ \hline
0.02 & 1.85 (0.58) & 1.94 (0.55) & 0.85 (0.38) & 0.87 (0.37) & 0.45 (0.22) & 0.41 (0.21) \\ \hline
0.04 & 4.91 (1.64) & 4.93 (0.73) & 2.34 (0.82) & 2.33 (0.51) & 1.14 (0.58) & 0.96 (0.36) \\ \hline
0.05 & 8.31 (5.49) & 7.66 (1.71) & 3.59 (1.05) & 3.64 (0.77) & 1.84 (0.88) & 1.52 (0.5) \\ \hline
 \end{tabular}
\caption{Training setting: inviscid. Test setting: viscous. The numbers report relative errors ($\times 10^{-2}$) of RL-WENO and WENO with the means and standard deviations of the errors among 25 random test initial conditions. We again observe that RL-WENO can generalize to this new test setting and achieves similar performance to WENO.}
\label{tab:generalize_to_viscous}
\end{table}

\subsection{Viscous Burgers Equation with forcing and fixed initial value}

\begin{table}[t]
\centering
\begin{tabular}{|c|c|c|c|c|}
\hline
$\eta$                & $\Delta x$ & RL-WENO & WENO & L3D \\ \hline
\multirow{3}{*}{0.01} & 0.02       & 4.64 (0.7) & 4.74 (0.7) & 3.09 (0.9) \\ \cline{2-5} 
                      & 0.04       & 10.36 (1.45) & 10.3 (1.35) & nan (nan) \\ \cline{2-5} 
                      & 0.05      & 13.33 (2.05) & 13.55 (2.07) & nan (nan)    \\ \hline
\multirow{3}{*}{0.02} & 0.02       & 2.41 (0.43) & 2.59 (0.45) & 6.89 (0.74)     \\ \cline{2-5} 
                      & 0.04       & 6.53 (1.13) & 6.6 (0.95) & 4.08 (1.12)    \\ \cline{2-5} 
                      & 0.05       & 9.19 (1.54) & 9.33 (1.4) & 7.58 (1.11)  \\ \hline
\multirow{3}{*}{0.04} & 0.02       & 1.09 (0.21) & 1.02 (0.27) & 11.01 (1.3)     \\ \cline{2-5} 
                      & 0.04       & 3.42 (0.65) & 3.68 (0.62) & 11.22 (1.22)    \\ \cline{2-5} 
                      & 0.05       & 4.97 (0.91) & 5.42 (0.81) & 9.37 (1.21)     \\ \hline
\end{tabular}
\caption{Training setting: forcing. Test setting: forcing. Comparison of RL-WENO, WENO and L3D with different viscosity and grid sizes. $\Delta t$ is computed as $0.1 \times \Delta x$. The numbers report relative errors ($\times 10^{-2}$) of RL-WENO, WENO and L3D with the means and standard deviations of the errors among 25 random test forcings.  `nan' means that the trained model diverges and the solution explodes when solving the equation. }
\label{tab:compare_l3d}
\end{table}

\para{Comparision between WENO and L3D.}
In this section, we train RL-WENO in the forcing setting, and compare it with WENO and a recently proposed supervise learning-based method: \textbf{L3D}~\cite{bar2019learning}. Note that this setting we use (Eq. ~\eqref{eq:l3d_condition}) is exactly the same as the training/testing setting used in L3D.

For L3D, we use the code released by the authors\footnote{https://github.com/google/data-driven-discretization-1d}, and train a model using their provided dataset. 
After training, we test the trained model on different grid sizes and viscosity.

The test results are shown in Table~\ref{tab:compare_l3d}. We observe that in three cases, L3D reports lower error than WENO and RL-WENO, and in all other test cases, L3D reports higher error than RL-WENO and WENO. Notably, the trained model by L3D diverges in two of the test cases, where the solution blows up during the evolution of the equation. This shows that L3D has worse generalization compared with RL-WENO.

\para{Out-of-distribution generalization to the other two settings. }

\begin{table}[t]
\scriptsize
\centering
\begin{tabular}{|c|c|c|c|c|c|c|c|c|}
\hline
\multirow{2}{*}{\diagbox{$\Delta x$}{$\eta$}} & \multicolumn{2}{c|}{0} & \multicolumn{2}{c|}{0.01} & \multicolumn{2}{c|}{0.02} & \multicolumn{2}{c|}{0.04} \\ \cline{2-9} 
   & RL-WENO         & WENO  & RL-WENO         & WENO         & RL-WENO         & WENO         & RL-WENO         & WENO         \\ \hline
0.02 & 13.06 (3.22) & 11.13 (3.83) & 1.9 (0.58) & 1.94 (0.55) & 0.92 (0.37) & 0.87 (0.37) & 0.49 (0.22) & 0.41 (0.21) \\ \hline
0.04 & 18.7 (7.65) & 19.48 (9.33) & 4.76 (1.19) & 4.93 (0.73) & 2.3 (0.78) & 2.33 (0.51) & 1.13 (0.62) & 0.96 (0.36) \\ \hline
0.05 & 28.57 (22.3) & 35.88 (27.48) & 7.75 (2.76) & 7.66 (1.71) & 3.51 (1.24) & 3.64 (0.77) & 1.72 (0.95) & 1.52 (0.5) \\ \hline
 \end{tabular}
\caption{Training setting: forcing. Test setting: inviscid and viscous. The numbers report relative errors ($\times 10^{-2}$) of RL-WENO and WENO with the means and standard deviations of the errors among 25 random test initial conditions. }
\label{tab:forcing-rl-weno-on-viscous-and-inviscid}
\end{table}

The generalization results of RL-WENO to the inviscid and viscous settings are shown in Table~\ref{tab:forcing-rl-weno-on-viscous-and-inviscid}. We again see that RL-WENO generalizes relatively well.
 We tried to train L3D in the inviscid/viscous settings, but could not get any decent models. Besides, the L3D model trained under the forcing setting cannot generalize to other settings (the model blows up during the evolution of the equation). In contrast, RL-WENO has much stronger flexibility and generalization: decent policies can be obtained when trained under various settings, and the model trained under one setting can generalize relatively well to other settings.  Table~\ref{tab:experiment_summary} gives a thorough discussion of the generalization ability of L3D and RL-WENO.



\subsection{Viscous Burgers Equation with random initial conditions}

\begin{table}[t]
\scriptsize
\centering
\begin{tabular}{|c|c|c|c|c|c|c|c|c|}
\hline
\multirow{2}{*}{\diagbox{$\Delta x$}{$\eta$}} & \multicolumn{2}{c|}{0} & \multicolumn{2}{c|}{0.01} & \multicolumn{2}{c|}{0.02} & \multicolumn{2}{c|}{0.04} \\ \cline{2-9} 
   & RL-WENO         & WENO  & RL-WENO         & WENO         & RL-WENO         & WENO         & RL-WENO         & WENO         \\ \hline
0.02 & 11.17 (3.8) & 11.13 (3.83) & 1.64 (0.57) & 1.94 (0.55) & 0.74 (0.35) & 0.87 (0.37) & 0.42 (0.21) & 0.41 (0.21) \\ \hline
0.04 & 19.37 (11.67) & 19.48 (9.33) & 4.29 (0.93) & 4.93 (0.73) & 1.91 (0.58) & 2.33 (0.51) & 0.87 (0.38) & 0.96 (0.36) \\ \hline
0.05 & 41.44 (45.05) & 35.88 (27.48) & 7.6 (3.94) & 7.66 (1.71) & 2.97 (1.07) & 3.64 (0.77) & 1.38 (0.61) & 1.52 (0.5) \\ \hline
 \end{tabular}
\caption{Training setting: viscous. Test setting: inviscid/viscous. The numbers report relative errors ($\times 10^{-2}$) of RL-WENO and WENO with the means and standard deviations of the errors among 25 random test initial conditions.}
\label{tab:viscous-rl-weno-on-viscous-and-inviscid}
\end{table}

\para{Comparison between WENO and PINNs. }
The test results of RL-WENO compared with WENO are shown in the last 3 columns of Table~\ref{tab:viscous-rl-weno-on-viscous-and-inviscid}. We again see that in most cases, RL-WENO outperforms WENO.

The comparison results of RL-WENO versus PINNs are shown in Table~\ref{tab:compare_PINNs}. Again, we see that PINNs performs poorly, due to the complexity of our chosen initial values. 

\para{Out-of-distribution generalization to the other two test settings. }

\begin{table}[t]
\footnotesize
\centering
\begin{tabular}{|c|c|c|c|c|c|c|}
\hline
\multirow{2}{*}{\diagbox{$\Delta x$}{$\eta$}}  & \multicolumn{2}{c|}{0.01} & \multicolumn{2}{c|}{0.02} & \multicolumn{2}{c|}{0.04} \\ \cline{2-7} 
 & RL-WENO         & WENO         & RL-WENO         & WENO         & RL-WENO         & WENO         \\ \hline
0.01 & 4.66 (0.72) & 4.74 (0.7) & 2.33 (0.47) & 2.59 (0.45) & 1.01 (0.19) & 1.02 (0.27) \\\hline
0.02 & 10.28 (1.33) & 10.3 (1.35) & 6.43 (0.88) & 6.6 (0.95) & 3.43 (0.61) & 3.68 (0.62) \\\hline
0.04 & 13.38 (2.3) & 13.55 (2.07) & 9.08 (1.57) & 9.33 (1.4) & 5.19 (0.92) & 5.42 (0.81) \\\hline
 \end{tabular}
\caption{Training setting: viscous. Test setting: forcing. The numbers report relative errors ($\times 10^{-2}$) of RL-WENO and WENO with the means and standard deviations of the errors among 25 random test forcings.}
\label{tab:viscous-rl-weno-on-forcing}
\end{table}

The generalization results to the inviscid version is shown in the first column of Table~\ref{tab:viscous-rl-weno-on-viscous-and-inviscid}, and the generalization results to the forcing setting is shown in Table~\ref{tab:viscous-rl-weno-on-forcing}. We again see that RL-WENO  generalizes relatively well to these new settings.


%% file: 05_conclusion.tex
\section{Conclusion and Future Work}
\label{sec::conclusion}

In this paper, we proposed a general framework to learn how to solve 1-dimensional conservation laws via deep reinforcement learning. We first discussed how the procedure of numerically solving conservation laws can be naturally cast in the form of Markov Decision Process. We then elaborated how to relate notions in numerical schemes of PDEs with those of reinforcement learning. In particular, we introduced a numerical flux policy which was able to decide on how numerical flux should be designed locally based on the current state of the solution. 
We carefully design the action of our RL policy to 
make it a meta-learner. 
Our numerical experiments showed that the proposed RL based solver was able to outperform high order WENO, which is one of the most delicate and successful numerical solvers for conservation laws. Furthermore, RL based solver has notably better generalization than the supervised learning based approach L3D and PINNs, especially for out-of distribution generalization, i.e. when the test settings are vastly different from the training settings.

As part of the future works, we would like to consider using the numerical flux policy to inference more complicated numerical fluxes with guaranteed consistency and stability. Furthermore, we can use the proposed framework to learn a policy that can generate adaptive grids and the associated numerical schemes. Lastly, we would like consider system of conservation laws in 2nd and 3rd dimensional space.

\section*{Acknowledgments}
Bin Dong is supported in part by National Natural Science Foundation of China (NSFC) grant No. 11831002, Beijing Natural Science Foundation (No. 180001) and Beijing Academy of Artificial Intelligence (BAAI).